\documentclass{article}
\usepackage{ijcai24}

\usepackage{times}
\usepackage{soul}
\usepackage{url}
\usepackage[hidelinks]{hyperref}
\usepackage[utf8]{inputenc}
\usepackage[small]{caption}
\usepackage{graphicx}
\usepackage{amsmath}
\usepackage{amsthm}
\usepackage{booktabs}
\usepackage{algorithm}
\usepackage{algorithmic}
\usepackage[switch]{lineno}

\usepackage{authblk}
\usepackage{natbib}
\usepackage{subfigure}
\usepackage{amssymb}
\usepackage{mathtools}
\usepackage{amsthm}
\usepackage{amsfonts, dsfont}
\usepackage{makecell}
\usepackage{arydshln}
\usepackage{multirow}
\usepackage{bm}
\usepackage{xcolor}
\usepackage{listings}
\usepackage{booktabs} 
\newcolumntype{x}[1]{>{\centering\arraybackslash}p{#1}}

\title{Momentum Boosted Episodic Memory for Improving Learning in Long-Tailed RL Environments}

\author[1]{Dolton Fernandes\thanks{These authors contributed equally to this work.}}
\author[1,2]{Pramod Kaushik}
\author[1]{Harsh Shukla}
\author[1]{Bapi Raju Surampudi}

\affil[1]{IIIT Hyderabad}
\affil[2]{TCS Research, Pune}

\begin{document}

\maketitle

\begin{abstract}
Traditional Reinforcement Learning (RL) algorithms assume the distribution of the data to be uniform or mostly uniform. However, this is not the case with most real-world applications like autonomous driving or in nature where animals roam. Some experiences are encountered frequently, and most of the remaining experiences occur rarely; the resulting distribution is called \emph{Zipfian}. Taking inspiration from the theory of \emph{complementary learning systems}, an architecture for learning from Zipfian distributions is proposed where important long tail trajectories are discovered in an unsupervised manner. The proposal comprises an episodic memory buffer containing a prioritised memory module to ensure important rare trajectories are kept longer to address the Zipfian problem, which needs credit assignment to happen in a sample efficient manner. The experiences are then reinstated from episodic memory and given weighted importance forming the trajectory to be executed. Notably, the proposed architecture is modular, can be incorporated in any RL architecture and yields improved performance in multiple Zipfian tasks over traditional architectures. Our method outperforms IMPALA by a significant margin on all three tasks and all three evaluation metrics (Zipfian, Uniform, and Rare Accuracy) and also gives improvements on most Atari environments that are considered challenging.

\end{abstract}

\section{Introduction}

Humans and animals roam around in environments that are unstructured in nature. However, existing algorithms in reinforcement learning are built around the assumption that environments are mostly uniform. Most of the time, a small subset of experiences frequently recur while many important experiences occur only rarely ~\citep{zipf2013psycho, smith2018developing}. For example, consider a deer attempting to survive in a habitat with predators. If the deer successfully evades a potential predator while drinking water from a source, it cannot afford to gradually learn from multiple experiences of this nature in order to avoid hazardous locations associated with water sources. Instead, the deer must rapidly learn from this experience and generalize it to similar situations. Similarly, in autonomous driving, experiences are not uniform, and usually, the rare instances where there is an accident or unusual experiences are more critical in real-world settings. This is the fundamental premise on which the theory of complementary learning systems ~\citep{mcclelland1995there, kumaran2016learning} is proposed. In this framework, an intelligent agent needs to have a fast learning system and a slow learning system operating together to restructure the statistics of the environment for better survival internally and not be naive by expecting uniform environments. In the brain, this is hypothesized through the interplay between the hippocampus, a fast learning system, and the neocortex which is a slow learning system, and together they manage to generalize and retain experiences crucial to the goals of the organism ~\citep{kumaran2016learning, botvinick2019reinforcement}. 
The hippocampus is able to achieve fast learning through its reliance on the slow learning system of the cortex where high dimensional data coming from the sensory systems are converted to low dimensional representations which can be operated on by the hippocampus. Such top-down modulation from the cortex influences the processing in the hippocampus ~\citep{kumaran2007computational}. Similarly, the slow structured learning of the cortex happens through interleaved learning by replaying experiences stored in the hippocampus ~\citep{o2010play}.

Here, we particularly look at this interplay between a fast learning and a slow learning system and apply this to solve the long-tailed phenomena. The reinforcement learning algorithm ~\citep{sutton2018reinforcement,espeholt2018impala} uses the episodic buffer to generalize across experiences, and a familiarity memory prioritizes long-tail data from the outputs generated by the RL algorithm. This prioritization of samples happens through a contrastive learning-related momentum loss which enables the unsupervised discovery of long-tailed data from the stream of experiences ~\citep{zhou2022contrastive}. These prioritized samples are retained for a longer duration in memory so that the corresponding hidden activations may be reinstated in the recurrent layers of the RL network.


Our main contributions are:
\begin{itemize}
\item Proposing a first solution for the problem of navigating to objects occurring with a long tail distribution using deep reinforcement learning. 
\item Application of contrastive momentum loss for unsupervised discovery of long tail states in the context of reinforcement learning.
\item Novel method to prioritize long-tail states in the buffer then reinstating hidden activations in recurrent layers.
\end{itemize}

\section{Background}

\subsection{Markov Decision Processes}
Let's assume an environment $\mathcal{E}$ which provides the agent with an observation $S_t$, the agent selects an action $A_t$, and then the environment responds by providing the agent with the next state $S_{t+1}$. The interactions between the agent and environment are formalized by \textit{MDPs} which are reinforcement learning tasks that satisfy Markov property ~\citep{SuttonBarto}. It is defined by the tuple $<\mathcal{S}, \mathcal{A}, \mathcal{R}, \mathcal{T}, \gamma >$ where, $\mathcal{S}$ represents the set of states, $\mathcal{A}$ is the set of actions, $\mathcal{R}: \mathcal{S} \times \mathcal{A} \to \mathbb{R}$ denotes the reward function. $\mathcal{T}: \mathcal{S} \times \mathcal{A} \to Dist(\mathcal{S})$ represents the transition function mapping state-action pairs to a distribution over next states $Dist(\mathcal{S})$ and $\gamma \in [0,1]$ is the discount factor. We consider IMPALA as our base architecture and build on top of it ~\citep{espeholt2018impala}.

\subsection{Memory Systems in RL}

 Memory systems in humans allow them to retrieve the relevant set of experiences for decision-making in case of unseen circumstances. In neuroscience, some of the types of memories studied are - \textit{Working Memory} and \textit{Episodic Memory}. Working memory is short-term temporary storage while episodic memory is a non-parametric or semi-parametric long-term storage memory. Deep Reinforcement Learning agents with episodic memory, in particular, a combination of non-parametric and parametric networks have shown improved sample efficiency and are suitable for decision-making in rare events.~\cite{Blundell} used a non-parametric model to keep the best Q-values in tabular memory. \cite{Pritzel_Paper} in Neural Episodic Control proposed a differentiable-neural-dictionary to keep the representations and Q-values in a semi-tabular form. \cite{Hansen} undertook a trajectory-centric approach to model such systems. Our approach adopts a unique perspective on Episodic Memory (MEM), integrating a latent recurrent neural network for working memory functionality alongside an episodic memory component. This methodology, while distinctive in its implementation, parallels certain conceptual frameworks in the realm of neural network-based memory systems \cite{fortunato2019generalization}.

\begin{figure*}[ht!]
\begin{center}
\includegraphics[width=0.65\textwidth]{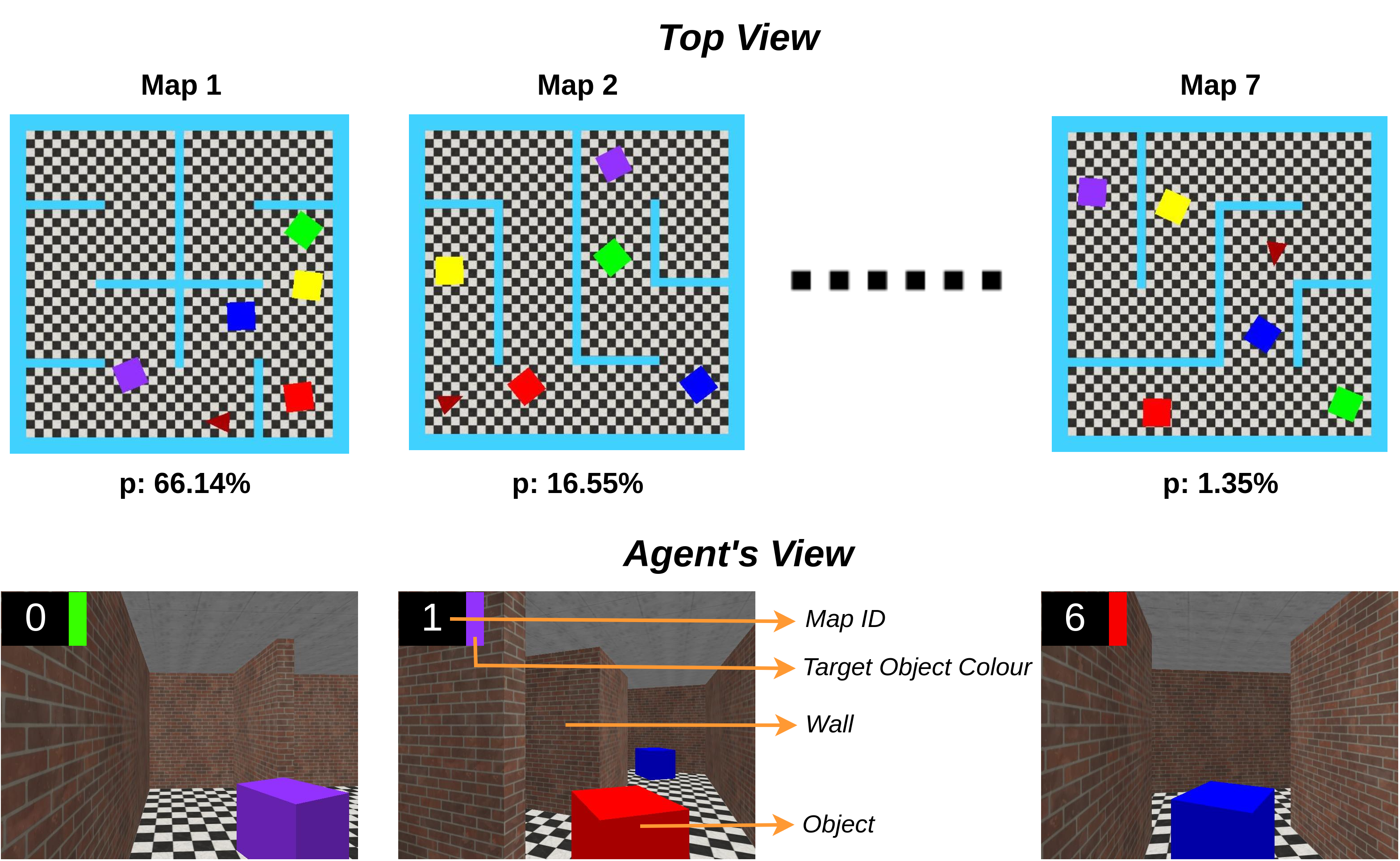}
\end{center}
\vspace{-0.4cm}
\caption{\textbf{Zipf's 3DWorld Task:} Contains 7 maps, each with 5 objects placed at random locations. The location of these objects does not change during trials. The agent (Red triangle in top view) starts at a fixed location in each trial and has to navigate towards the target object, whose color is shown in the top-left corner along with the current map ID (0 indexed). The agent's first-person view of each map is shown in the bottom images. The details of the environment experienced can be seen in the annotated image. The value 'p' below each map shows the probability of occurrence of the map in a trial, highlighting the skew in the distribution. A similar skew occurs for the distribution of objects in these maps. We can see this in Figure~\ref{fig:1b}, which shows the distribution of objects for the first map (most common).}\label{fig:1a}
\end{figure*}

\begin{figure}[ht]
\centering
\includegraphics[width=0.4\textwidth]{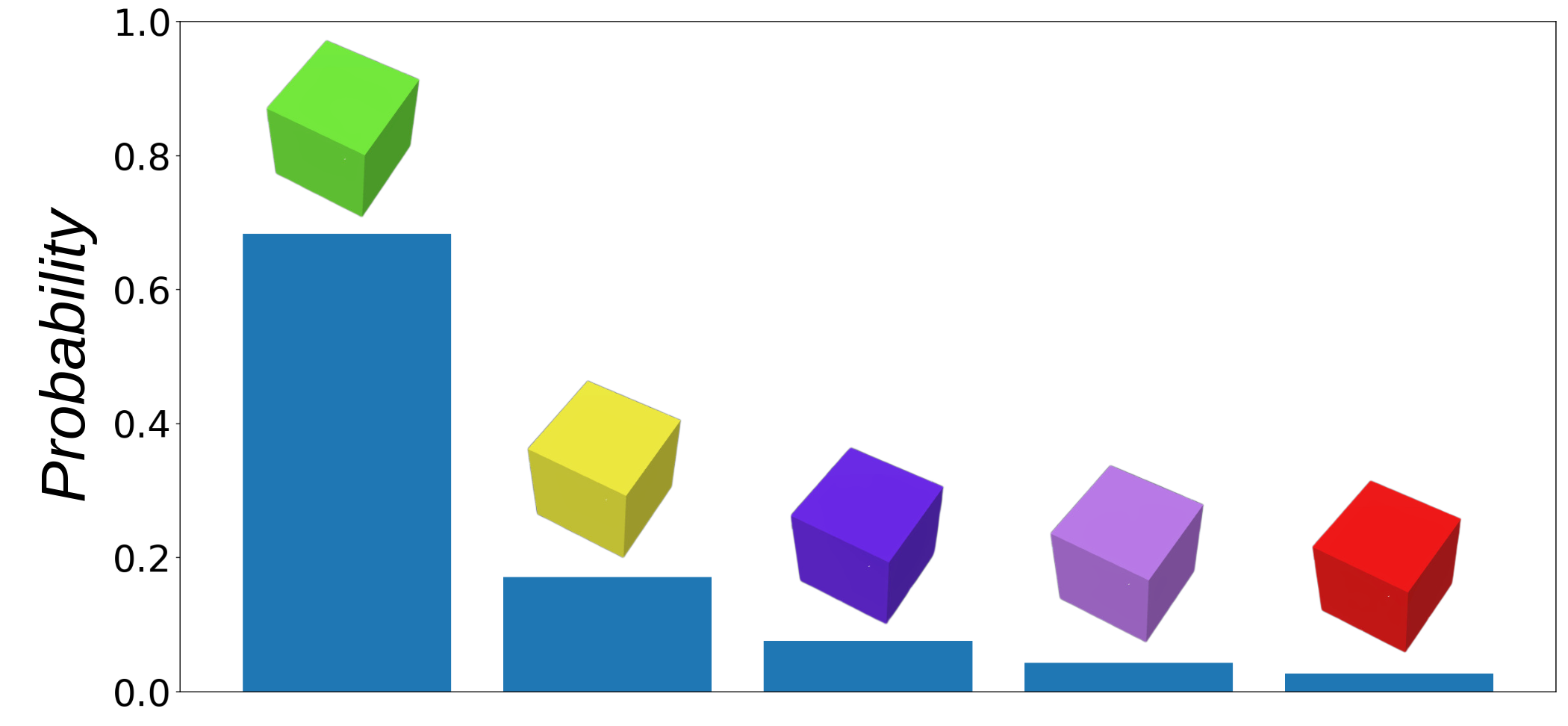}
\vspace{-0.3cm}
\caption{The probability distribution for objects to appear as the target object in a map during a trial. This example shows the distribution of objects for Map 1 in Figure ~
\ref{fig:1a}.}\label{fig:1b}
\end{figure}

\begin{figure*}[ht!]
\centering
\begin{tabular}{l}
\begin{minipage}[b]{1.0 \linewidth}
\begin{center}
\subfigure[Input]{
\includegraphics[width=0.9in]{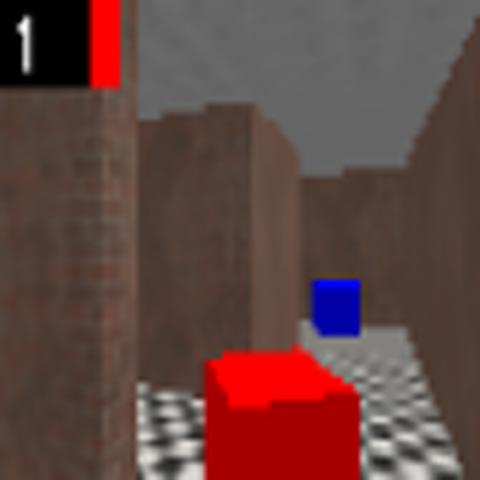}
\label{fig:2a}
}
\hspace{0.5cm}
\subfigure[Gaussian Noise]{
\includegraphics[width=0.9in]{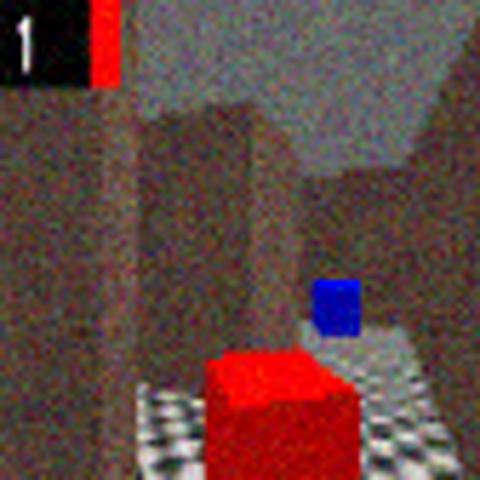}
\label{fig:2b}
}
\hspace{0.5cm}
\subfigure[Random Cutout]{
\includegraphics[width=0.9in]{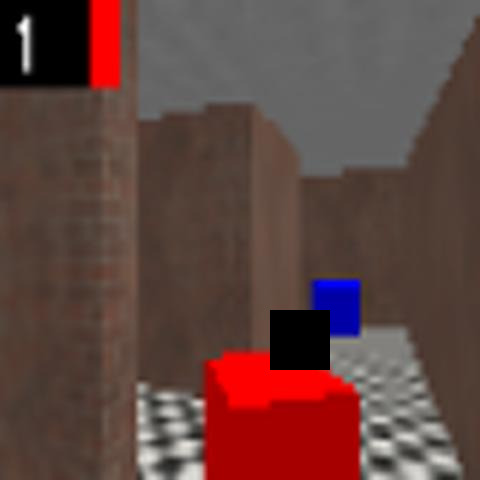}
\label{fig:2c}
}
\hspace{0.5cm}
\subfigure[Final Image]{
\includegraphics[width=0.9in]{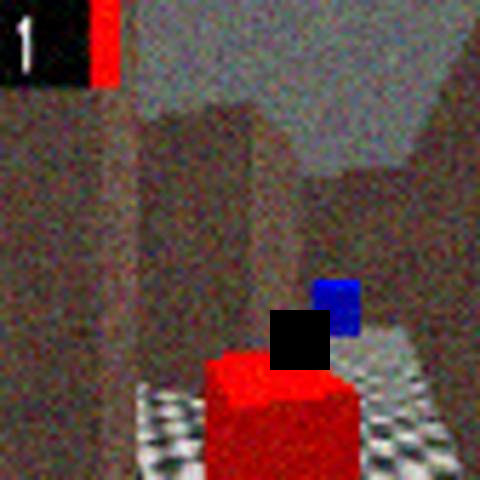}
\label{fig:2d}
}
\end{center}
\end{minipage}
\end{tabular}
\vspace{-0.5cm}
\caption{\textbf{Image augmentations for contrastive learning:} \textbf{(a)} Shows downsampled input image for a trial. \textbf{(b)} Input image after adding Gaussian noise to it. \textbf{(c)} Input image after applying random cutout augmentation. The black rectangle near the agent's position is the area cutout. \textbf{(d)} Final augmented image after adding Gaussian noise and random cutout.}
\label{fig:2}
\end{figure*}

\section{Environments}

We investigate the Zipf's Gridworld and Zipf's Labyrinth tasks introduced in ~\cite{Chan2022ZipfianEF}, which presents multiple distinct tasks consisting of skewed data distributions along various dimensions that challenge conventional architectures to generalize to rare states and events. We focus on tasks that only require visual input stimulus (images).

The Zipf's Gridworld task involves navigating to a target object in a partially observable gridworld environment. The positions of the objects in these maps do not change during trials. If the object selected at the end is the correct target object, the trial ends and the agent gets a positive reward. If the object selected is incorrect or the number of steps exceeds the limit, the agent gets no reward. We conduct our experiments on (10 maps, 10 objects).

Zipf's Labyrinth task focuses on the heavily-skewed experience of tasks and situations that pertain to specific goals. In each episode, a task is sampled (based on Zipf's distribution - Equation ~\ref{eqn:zipfslaw}) from the DM-Lab benchmark which is a collection of tasks set in a 3D and first-person environment built on Quake 3 Arena\footnote{\url{https://github.com/id-Software/Quake-III-Arena}}. 

We additionally test our method on a new 3D environment `Zipf's 3DWorld', that we propose in this paper.

Further, we conduct experiments with tasks from the Atari Learning Environment ~\citep{bellemare2013arcade} to test the robustness of the proposed algorithm in general environments.

\subsection{Zipf's 3DWorld}

We propose a task that is similar to Zipf's Gridworld task in ~\citep{Chan2022ZipfianEF} but in a 3D setting. The task involves navigating to a target object in a partially observable 3D environment. There are a total of 7 maps, and in each of these maps, 5 objects are placed at random positions. In each map, the starting location of the agent and the characteristics of objects such as their shape, color, and location are fixed. Given a target object, the agent has to find the object in the partially observable environment by taking at most 200 steps. If the object picked at the end is the correct target object, the trial ends and the agent gets a positive reward. If the object selected is incorrect or the number of steps exceeds the limit, the agent gets no reward.

The target object during a trial is embedded in the top left corner of the visual input of the agent. Since all the objects in the environments are square boxes, only the target object color is shown. Along with it, to the left of the target object color, the current map ID is also embedded, as shown in Figure~\ref{fig:1a}. The actions (move forward, move backward, turn left,  turn right \& pick object) can be used to navigate to any place in the environment. It is ensured that all the objects in a map are distinct, and the agent is able to reach any object present on the map using the set of five actions. An example of the agent's partial view can be seen in Figure~\ref{fig:1a}.

\begin{equation}\label{eqn:zipfslaw}
zp(k, n, e)=\frac{1/k^e}{\sum_{i=1}^{n} (1/i^e)}
\end{equation}
The probability of occurrence of the maps is governed by Zipf's power law (Equation ~\ref{eqn:zipfslaw}), where $n$ is the number of maps, $k$ is the map index $(1 <= k <= n)$ and $e$ is the Zipf's exponent. The same skew can be seen for target object selection in each map as shown in Figure~\ref{fig:1b}. To solve the task, the agent not only needs to explore the environment cleverly but also needs to memorize the path if the agent solves the trial correctly.

\section{Architecture}

Given an image observation ($im$), IMPALA's ~\citep{espeholt2018impala} feature extractor produces a pixel input embedding ($p$). This embedding is passed to an LSTM network with the hidden state ($h$) to generate a new hidden state, policy ($\pi$), and value ($V$). Our architecture includes a MEM module (red buffer in Figure~\ref{fig:4}) that stores state embeddings and associated LSTM hidden states (working memory) for identified rare or tail states. The MEM module is used to retrieve the relevant memory ($m$) for rare states, which is then additionally fed as input to the LSTM network along with the pixel input embedding $p$. We introduce an additional `familiarity' buffer (light cyan buffer in Figure~\ref{fig:4}) that employs boosted contrastive learning to prioritize and filter rare states for storage in the MEM. 
The caching of rare states is also influenced by the caching of LSTM hidden activations, which is contingent on the RL loss and, consequently, a part of the policy learning progress. In other words, only rare trajectories that are significant to the task are stored in the hidden activations, ensuring that the architecture does not fall prey to the `noisy TV problem', which involves detecting pure novelty regardless of its relevance to the policy. 
In addition, the architecture follows a modular approach and can be integrated with any other RL architecture to improve its long tail performance.

\subsection{State familiarity using Boosted Contrastive Learning}

A `familiarity' buffer is a circular buffer that contains states that are processed to determine level of rarity. 
To achieve this, we take inspiration from ~\cite{zhou2022contrastive}, where they improve performance on a long-tailed self-supervised learning task by proposing a momentum loss that can predict which samples among the dataset are long-tail samples. 

\begin{figure*}[ht]
\centering
\includegraphics[width=0.65\textwidth]{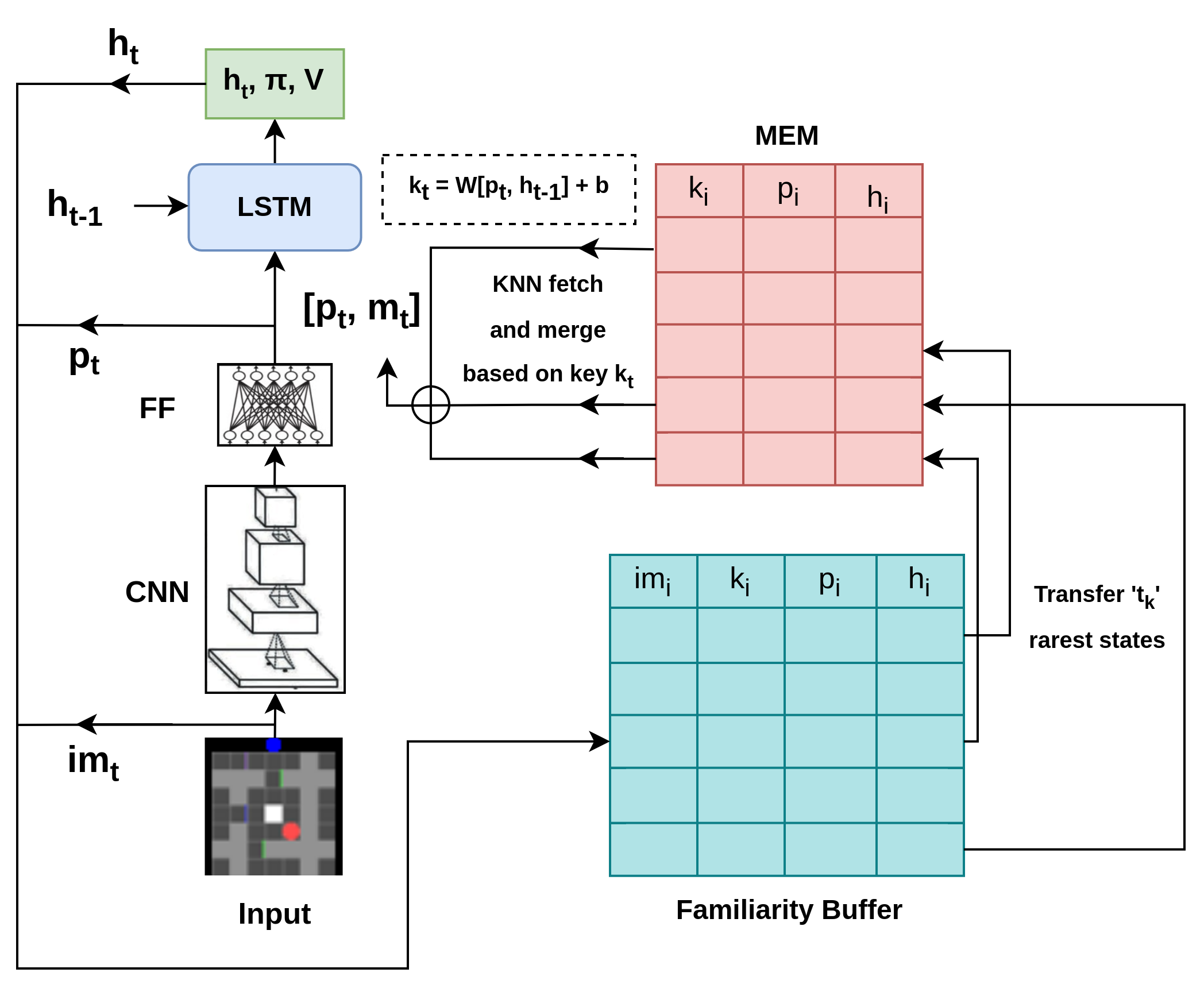}
\vspace{-0.3cm}
\caption{\textbf{Model Architecture:} The figure shows our momentum-boosted episodic memory architecture pipeline. The IMPALA backbone consists of a CNN feature extractor followed by a Feed Forward layer that gives the embedding. This embedding is concatenated with the one hot action encoding, reward \& memory to get pixel embedding  $p_i$ and then given to the LSTM network for further processing with working memory. The LSTM network additionally takes the past hidden state $h_{t-1}$ as input. During training the input image, pixel embedding, LSTM hidden states and keys are stored in the familiarity buffer. The momentum loss tracked on this buffer during contrastive learning is then used to prioritize long-tail states. The MEM is then periodically updated with top $t_f$ states from the familiarity buffer. The memory ($m_t$) is computed from the MEM using a weighted sum ($\bigoplus$) over results from a KNN similarity search on the keys present in the MEM using the query key $k_t$ (Equation ~\ref{eqn:wsum}).}\label{fig:4}
\end{figure*}

Our `familiarity' buffer (light cyan buffer in Figure~\ref{fig:4}) contains the input image ($im$), key ($k$), pixel input embedding ($p$), and LSTM hidden state ($h$). We will define these terms in the following section. In  each learning step of IMPALA ~\citep{espeholt2018impala}, a batch of trajectories is sent to this buffer. 

The feature extractor of IMPALA is trained using an additional auxiliary contrastive loss (Equation ~\ref{eqn:contrloss}) on states present in the familiarity buffer. Once the circular buffer is full, the contrastive learning process starts to help the policy learning process. For each image sample $im_i$ in the buffer, we find its pixel input embedding $p_i$ generated by IMPALA's feature extractor. The same feature extractor is used on augmented images to obtain the augmented embedding $p^{aug}_i$. The image $im_i$ is augmented using two augmentations, namely Gaussian noise ~\citep{boyat2015review} and random cutout ~\citep{devries2017improved} as shown in Figure~\ref{fig:2}. We use these augmentations because they are simple and have minimal impact on the task. 

For adding Gaussian noise (Figure~\ref{fig:2b}) to the image, we first generate an image of the same dimension as that of the original image,  filled with random numbers from $N(0, \sigma^2)$. This image is added to the original image after amplifying by a factor of $\sigma$. For random cutout (Figure~\ref{fig:2c}), we take a random location in the image and cut out a rectangular area of random size, replacing the pixels in the rectangular region with black pixels.

During the training process, we consider the embeddings $p_i$ and $p^{aug}_i$ as positive pairs for contrastive learning. The NT-Xent loss (~\cite{sohn2016improved}, Equation ~\ref{eqn:contrloss}) is used to calculate the loss per sample. For each sample $i$ in the familiarity buffer, we track its momentum loss following ~\cite{zhou2022contrastive}. The momentum loss helps to determine which samples in the familiarity buffer are long-tail samples. For a sample $i$ over $T$ consecutive epochs the contrastive losses are $\{\ell_{i,1}^T, \ell_{i,2}^T, ..., \ell_{i,T}^T\}$ and the moving average momentum loss is defined as follows:
\begin{equation}\label{eqn:momentum}
\ell m^m_{i,1} = \ell_{i,1}^T;~~\ell m^m_{i,t} = \beta \ell m^m_{i,t-1} + (1-\beta)\ell_{i,t}^T
\end{equation}
where $\beta$ is a hyperparameter that controls the degree smoothed by the historical losses. The final normalized momentum used to determine the familiarity of states is defined as
\begin{equation}\label{eqn:familiarity}
M_{i,t} = \frac{1}{2}\left(\frac{\ell m^m_{i,t}-{\bar{\ell m}^m_t}}{max\{|\ell m^m_{j,t}-{\bar{\ell m}^m_t}|\}_{j=1,{\dots},N}}+1\right)
\end{equation}
where $\bar{\ell m}^m_t$ is the average momentum loss of the dataset at the $t^{th}$ training step of the algorithm and N is the number of samples. The higher the momentum value $M_{i,t}$, the higher the rareness of the sample in the familiarity buffer. The model is trained end to end by optimizing both IMPALA's loss and the auxiliary contrastive loss. Let the loss given by IMPALA be $\mathcal{L}_{\mathrm{impala}}$ and that given by the contrastive learning branch be $\mathcal{L}_{\mathrm{contrastive}}$, then we define the final loss as shown in Equation ~\ref{eqn:finalloss} below.
\begin{equation}\label{eqn:finalloss}
\mathcal{L} = \mathcal{L}_{\mathrm{impala}} + \gamma*\mathcal{L}_{\mathrm{contrastive}}
\end{equation}
where $\gamma$ is a hyperparameter. The contrastive loss is given by:
\begin{equation}\label{eqn:contrloss}
\mathcal{L}_{\mathrm{contrastive}}=\frac{1}{N} \sum_{i=1}^{N}-\log \frac{\exp \left(\frac{p_{i}^\top.p^{aug}_{i}}{\tau}\right)}{\sum_{p_{i}' \in X^{'}} \exp \left(\frac{p_{i}^\top.p_{i}'}{\tau}\right)}
\end{equation}
where N is the number of samples, $X^{'}$ represents $X^{-}\cup \{p^{aug}_i\}$, $\left(p_i, p^{aug}_{i}\right)$ is the positive sample pair, $X^{-}$ is the negative sample set of $p$ and $\tau$ is the temperature.

\subsection{Combining Familiarity with Episodic Memory}

In this paper, an Episodic Memory (MEM) is also introduced on top of the IMPALA architecture, which is a circular buffer that stores the pixel embedding ($p$), LSTM hidden state ($h$), and the key ($k$). The key, $k$, is calculated using
\begin{equation}\label{eqn:multiply}
k = W[p,h]+b
\end{equation}
where W and b are learnable parameters and $[p,h]$ denotes the concatenation of $p$ and $h$ along the dimension axis. 
The MEM-enhanced IMPALA is designed to save summaries of previous experiences for the purpose of extracting crucial data that may be exploited by new states with a similar context. This is achieved by passing a summarised context (memory) to the main controller or agent, modifying decision making for the new state. Learning long-term dependencies, which can be challenging when depending solely on backpropagation in recurrent architectures, is made simpler by successfully enhancing the controller's working memory capacity with experiences from various time scales received from the MEM. 

In this algorithm, the pixel embeddings $p_i$, LSTM hidden states $h_i$, and keys $k_i$ are added to the familiarity buffer, which computes the familiarity of states based on the momentum loss and periodically sends updates to the MEM buffer to facilitate learning during rare trials. 
We only maintain relatively rare states in the MEM buffer to help learn about those rare states that actually require the help of an external episodic memory module. The MEM gets $t_k$ most rare states from the familiarity buffer after every $t_f$ training epochs of contrastive learning, where $t_k$ and $t_f$ are hyperparameters. Frequent states are handled as normal by the original IMPALA. 

\begin{algorithm}[h]
\caption{Pseudocode for our algorithm}\label{alg:alg1}
\textbf{Inputs:} Familiarity memory $\bm{fm}$, MEM $\bm{mem}$, transfer frequency $\bm{t_f}$, number of rare instances to transfer $\bm{t_k}$, number of IMPALA training epochs $\bm{T}$ and the contrastive loss weight $\bm{\gamma}$.\\
\textbf{Initialize:} $fm.buffer$, $mem.buffer \gets \{\}$\hfill $\triangleright$ clear buffers

\begin{algorithmic}
\FOR{$t$ in $1,\cdots,T$} 
\STATE $tr \gets get\_impala\_batch(t)$\hfill $\triangleright$ Get trajectories
\STATE $im, k, p, h, impala\_loss \gets train(tr, mem)$
\STATE $fm.add(im, k, p, h)$\hfill $\triangleright$ Trajectory added
\STATE $cl \gets fm.contr\_train()$\hfill $\triangleright$ Contr. Loss - Equation ~\ref{eqn:contrloss}
\IF{$t \mod t_f = 0$}
\STATE $mv \gets fm.calculate\_normalized\_momentum()$\\\hfill $\triangleright$ Equation ~\ref{eqn:familiarity}
\STATE $rare\_experiences \gets fm.get\_rare\_k(t_k, mv)$
\STATE $mem.add(rare\_experiences)$
\ENDIF
\STATE $final\_loss \gets impala\_loss + \gamma*cl$\hfill $\triangleright$ Equation ~\ref{eqn:finalloss}
\STATE $final\_loss.backward()$\hfill $\triangleright$ Backpropogate loss
\ENDFOR
\end{algorithmic}
\end{algorithm}

The overall architecture can be seen in Figure~\ref{fig:4}. For a stimulus $p_t$ and previous hidden state $h_{t-1}$, the agent chooses the most pertinent events to provide as input $m_t$ to the LSTM network using a type of dot-product attention ~\citep{DBLP:journals/corr/BahdanauCB14} over its MEM. Using the key $k_t$, formed by $p_t$ and $h_{t-1}$ using Equation ~\ref{eqn:multiply}, a K Nearest Neighbour (KNN) search is done over the keys in MEM to find the most relevant $K$ keys. The hidden states for these $K$ relevant items are combined using the below-weighted sum (Equation ~\ref{eqn:wsum}) to get additional input $m_t$ to be provided to the LSTM network.
\begin{equation}\label{eqn:wsum}
m_t = \frac{\sum_{i=1}^{K}w_ih_i}{\sum_{i=1}^{K}w_i}\;\;w_i = \frac{1}{||k_t-W[p_i,h_i]-b||^2_2+\epsilon}
\end{equation}
where $\epsilon$ is a small constant and $||x||^2_2$ represents the squared $L2$ norm of $x$.

The pseudo-code for our algorithm is provided in Algorithm ~\ref{alg:alg1}.

\section{Experiments and Results}

\begin{table*}[htb]
	\caption{\textbf{Evaluation Performance:} We compare four different methods with our algorithm namely IMPALA, IMPALA+Visual Reconstruction using a simple CNN-based autoencoder, IMPALA+MEM, and IMPALA+MEM with only contrastive learning. We report median results across three runs ($\pm$ absolute median deviation across runs) with distinct random seeds for models trained for $4 \times 10^7$ steps. Our method (IMPALA+MEM+Contrastive Learning+Rare State Prioritization using Familiarity Buffer) beats the remaining methods on all three evaluation metrics.}\label{table:performance}
	\vskip 0.15in
    \begin{minipage}{\linewidth}
		\resizebox{\linewidth}{!}{
		\begin{tabular}{x{0.441\linewidth}x{0.147\linewidth}x{0.147\linewidth}x{0.147\linewidth}}
			\toprule
            \multirow{2}{*}{\parbox{3.5cm}{\centering \phantom{} \\ \phantom{} \\ Method}} & \multicolumn{3}{c}{Accuracy (\%) $\mid$ \textbf{Zipf's Gridworld}}                                                          \\ \cmidrule(lr){2-4} 
										& \multicolumn{1}{c}{Zipfian} & \multicolumn{1}{c}{\parbox{3.5cm}{\centering Uniform\\(All maps and objects)}} & \multicolumn{1}{c}{\parbox{3.5cm}{\centering Rare\\(Rarest 20\% objects\\on rarest 20\% maps)}}   \\ \midrule
IMPALA                                &  88.3 $\pm$ 2.1 & \parbox{3.5cm}{\centering 41.1 $\pm$ 1.8} &  \parbox{3.5cm}{\centering 0.0 $\pm$ 0.0} \\
IMPALA + Visual Reconstruction        &  90.2 $\pm$ 2.4 & \parbox{3.5cm}{\centering 45.9 $\pm$ 1.1} &  \parbox{3.5cm}{\centering 0.0 $\pm$ 0.0} \\
IMPALA + MEM                          &  92.9 $\pm$ 3.2 & \parbox{3.5cm}{\centering 51.2 $\pm$ 2.3} & \parbox{3.5cm}{\centering 25.0 $\pm$ 2.0} \\
IMPALA + MEM + Contrastive Learning   &  94.8 $\pm$ 2.7 & \parbox{3.5cm}{\centering 52.3 $\pm$ 1.1} & \parbox{3.5cm}{\centering 25.1 $\pm$ 2.4} \\ \hdashline \\[-8pt]
\textbf{Ours}                         &  \textbf{98.5 $\pm$ 1.2} & \parbox{3.5cm}{\centering \textbf{66.3 $\pm$ 1.0}} & \parbox{3.5cm}{\centering \textbf{25.2 $\pm$ 1.1}} \\
			\bottomrule
        \end{tabular}
		}
	\end{minipage}\\[12pt]

    \begin{minipage}{\linewidth}
		\resizebox{\linewidth}{!}{
		\begin{tabular}{x{0.441\linewidth}x{0.147\linewidth}x{0.147\linewidth}x{0.147\linewidth}}
			\toprule
            \multirow{2}{*}{\parbox{3.5cm}{\centering \phantom{} \\ \phantom{} \\ Method}} & \multicolumn{3}{c}{Accuracy (\%) $\mid$ \textbf{Zipf's 3DWorld}}                                                          \\ \cmidrule(lr){2-4} 
										& \multicolumn{1}{c}{Zipfian} & \multicolumn{1}{c}{\parbox{3.5cm}{\centering Uniform\\(All maps and objects)}} & \multicolumn{1}{c}{\parbox{3.5cm}{\centering Rare\\(Rarest 20\% objects\\on rarest 20\% maps)}}   \\ \midrule
IMPALA                                &  95.9 $\pm$ 4.1 & \parbox{3.5cm}{\centering 65.6 $\pm$ 8.2} &  \parbox{3.5cm}{\centering 33.3 $\pm$ 2.1} \\
IMPALA + Visual Reconstruction        &  96.0 $\pm$ 2.2 & \parbox{3.5cm}{\centering 68.6 $\pm$ 10.4} &  \parbox{3.5cm}{\centering 37.1 $\pm$ 3.1} \\
IMPALA + MEM                          &  97.3 $\pm$ 2.2 & \parbox{3.5cm}{\centering 74.3 $\pm$ 6.0} & \parbox{3.5cm}{\centering 42.6 $\pm$ 8.3} \\
IMPALA + MEM + Contrastive Learning   &  97.5 $\pm$ 1.8 & \parbox{3.5cm}{\centering 74.4 $\pm$ 5.5} & \parbox{3.5cm}{\centering 43.0 $\pm$ 1.2} \\ \hdashline \\[-8pt]
\textbf{Ours}                         &  \textbf{99.2 $\pm$ 1.3} & \parbox{3.5cm}{\centering \textbf{80.1 $\pm$ 2.0}} & \parbox{3.5cm}{\centering \textbf{55.6 $\pm$ 2.4}} \\
			\bottomrule
        \end{tabular}
		}
	\end{minipage}\\[12pt]

    \begin{minipage}{\linewidth}
            \renewcommand{\arraystretch}{1.4}
		\resizebox{\linewidth}{!}{
		\begin{tabular}{x{0.35\linewidth}x{0.10\linewidth}x{0.10\linewidth}x{0.10\linewidth}x{0.10\linewidth}x{0.10\linewidth}x{0.10\linewidth}}
			\toprule
            \multirow{2}{*}{\parbox{3.5cm}{\centering \phantom{} \\ \phantom{} \\ \phantom{} \\ Method}} & \multicolumn{6}{c}{Accuracy (\%) $\mid$ \textbf{Zipf's Labyrinth}} \\ \cmidrule(lr){2-7}
                                            & \multicolumn{3}{c}{Forward Zipf} & \multicolumn{3}{c}{Reversed Zipf} \\ \cmidrule(lr){2-4} \cmidrule(lr){5-7}
										& \multicolumn{1}{c}{Zipfian}	& \multicolumn{1}{c}{\parbox{3.5cm}{\centering Uniform\\(All maps and objects)}}	& \multicolumn{1}{c}{\parbox{3.5cm}{\centering Rare\\(Rarest 20\% objects\\on rarest 20\% maps)}} & \multicolumn{1}{c}{\parbox{3.5cm}{\centering Zipfian}}	& \multicolumn{1}{c}{\parbox{3.5cm}{\centering Uniform\\(All maps and objects)}}	& \multicolumn{1}{c}{\parbox{3.5cm}{\centering Rare\\(Rarest 20\% objects\\on rarest 20\% maps)}}   \\ \midrule
IMPALA                                &  63.3 $\pm$ 3.0 & \parbox{3.5cm}{\centering 27.9 $\pm$ 3.1} &  \parbox{3.5cm}{\centering 5.0 $\pm$ 4.2} &  \parbox{3.5cm}{\centering 53.8 $\pm$ 7.1} &  \parbox{3.5cm}{\centering 21.3 $\pm$ 3.3} &  \parbox{3.5cm}{\centering 4.1 $\pm$ 3.1} \\
IMPALA + Visual Reconstruction                                &  65.6 $\pm$ 8.5 & \parbox{3.5cm}{\centering 31.2 $\pm$ 1.9} &  \parbox{3.5cm}{\centering 8.7 $\pm$ 4.2} &  \parbox{3.5cm}{\centering 55.6 $\pm$ 13.2} &  \parbox{3.5cm}{\centering 25.2 $\pm$ 5.1} &  \parbox{3.5cm}{\centering 9.6 $\pm$ 3.3} \\
IMPALA + MEM                                &  68.5 $\pm$ 7.6 & \parbox{3.5cm}{\centering 45.3 $\pm$ 2.5} &  \parbox{3.5cm}{\centering 23.1 $\pm$ 3.5} &  \parbox{3.5cm}{\centering 66.7 $\pm$ 11.3} &  \parbox{3.5cm}{\centering 38.6 $\pm$ 2.0} &  \parbox{3.5cm}{\centering 16.7 $\pm$ 2.7} \\
IMPALA + MEM + Contrastive Learning                                &  69.1 $\pm$ 4.6 & \parbox{3.5cm}{\centering 49.3 $\pm$ 4.4} &  \parbox{3.5cm}{\centering 27.5 $\pm$ 2.2} &  \parbox{3.5cm}{\centering 68.9 $\pm$ 9.4} &  \parbox{3.5cm}{\centering 39.0 $\pm$ 7.7} &  \parbox{3.5cm}{\centering 18.8 $\pm$ 5.1} \\ \hdashline \\[-8pt]
\textbf{Ours}                                &  \textbf{71.3 $\pm$ 3.1} & \parbox{3.5cm}{\centering \textbf{54.2 $\pm$ 2.2}} &  \parbox{3.5cm}{\centering \textbf{32.2 $\pm$ 2.1}} &  \parbox{3.5cm}{\centering \textbf{77.5 $\pm$ 7.0}} &  \parbox{3.5cm}{\centering \textbf{47.1 $\pm$ 2.3}} &  \parbox{3.5cm}{\centering \textbf{25.3 $\pm$ 1.9}} \\
			\bottomrule
        \end{tabular}
		}
	\end{minipage}\\
\vskip -0.1in
\end{table*}

\begin{figure*}
\begin{center}
\includegraphics[width=0.9\textwidth]{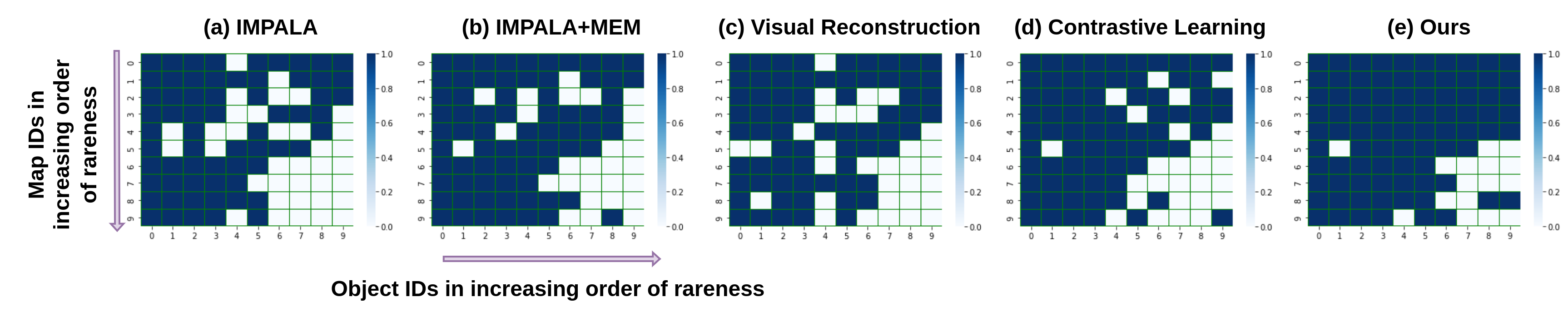}
\end{center}
\vspace{-0.7cm}
\caption{\textbf{Performance plots (Zipf's Gridworld): } \textbf{(a)} Performance of IMPALA agent on each map and object. The y-axis denotes the map axis, and the x-axis denotes the object axis. Value at (i, j) shows the performance (0-1 scale) of the agent on the trial where the object with ID j is chosen at the map with ID i. An increase in i and j means an increase in the rareness of the map and object respectively according to the Zipf's distribution (Equation ~\ref{eqn:zipfslaw}). \textbf{(b)} Performance of IMPALA with MEM added. We can see there are some medium-rare trials in which the agent has learned to navigate and learn the task. \textbf{(c)} IMPALA with Visual Reconstruction using CNN-based autoencoder. \textbf{(d)} Performance of IMPALA+MEM with contrastive learning. \textbf{(e)} Performance of our agent consisting of familiarity buffer that highlights long tail samples for MEM using modified boosted contrastive learning.}\label{fig:3}
\end{figure*}

We compare the results of our method with mainly four different types of architecture. The first architecture is the original IMPALA ~\citep{espeholt2018impala}, which is an off-policy actor-critic framework and has shown substantial improvements over baselines like ~\cite{clemente2017efficient, mnih2016asynchronous}. Second is  IMPALA with an episodic memory module. The third experiment is IMPALA with visual reconstruction. \cite{Chan2022ZipfianEF} have experimented with visual reconstruction ~\citep{DBLP:conf/iclr/HillTGWMC21}, and similarly we add an extra task for visual reconstruction on top of IMPALA with a CNN-based autoencoder. The fourth experiment includes contrastive learning to learn good embeddings. This approach does not find rare states in the familiarity buffer, however, but samples $k$ states uniformly randomly from the familiarity buffer instead. By contrast, in the proposed approach, we pass the rare $k$ states to MEM from the familiarity buffer.

From the experiments (Figure~\ref{fig:3} \& Table~\ref{table:performance}), we can see clearly that contrastive learning (feature representation learning) alone does not result in good performance and that we also need the familiarity buffer prioritizing rare states for the MEM. The training curves (Training/Zipfian Accuracy) and different ablations can be seen in the supplementary section. We observe that the Zipfian mean episode return for our method increases faster than all other methods in the initial phase of training and also converges later to achieve the highest accuracy. The hyperparameters used for our model across all tasks are listed in the supplementary section.

Having a higher training accuracy is not what we are looking for; instead, we want to achieve good accuracy when tested uniformly or in rare instances. Figures~\ref{fig:3}(a-e) show the performance of the five architectures on all the maps and objects of our environment for Zipf's Gridworld task. For each (map, object) combination, we plot the average performance across 50 trials. Figure~\ref{fig:3}a shows the performance of the IMPALA agent. We see that the agent is unable to learn extremely rare trials as well as some of the medium-rare trials. Figures~\ref{fig:3} (b) \& (d) show the performance of IMPALA+MEM with and without contrastive learning respectively. In the case of IMPALA+MEM with contrastive learning, the familiarity buffer is sampled uniformly randomly to fetch states for the MEM. We see that the performance is almost similar, however the medium-rare trials are not learned. Figure~\ref{fig:3}c shows the performance of IMPALA with added visual reconstruction using a CNN based autoencoder. This performs slightly better than the baseline (IMPALA) but fails to match the performance of other agents. Finally, in Figure~\ref{fig:3}e we see the performance of our agent with the familiarity module where medium-rare trials and some of the very rare trials are both successfully learned by our agent.

Table ~\ref{table:performance} gives more insight into our agent's performance on different tasks where our agent demonstrates significantly better performance compared to other agents on all three evaluation metrics (Training/Zipfian accuracy, Uniform accuracy, and Rare accuracy). We also tried using just the single value of contrastive loss instead of calculating the momentum loss over the history of contrastive losses, but the results were worse. This is because the momentum additionally captures the change in the contrastive losses, which predicts how fast and how well it is able to learn those samples. We also report results (see Supplementary) for the Variational Autoencoder (VAE) ~\citep{pu2016variational} \& Hierarchical Chunk Attention Memory (HCAM) ~\cite{lampinen2021towards} which have very different methods of solving and memory.

Results of additional experiments based on the Atari Learning Environment ~\cite{bellemare2013arcade} are in the supplementary section. The comparison is made with two other architectures namely, IMPALA (shallow) and IMPALA (deep), where the latter has a deeper perceptual processing module as compared to both the former approach and our proposed method. Our method outperforms both variations of IMPALA on 32 out of the 56 tasks considered (57.14\%) (Table ~\ref{table:atari_individual_games}). We obtain better results on 7/10 tasks mentioned in the \emph{challenging set} that emphasizes hard exploration tasks with long-term credit assignment ~\citep{pmlr-v119-badia20a} (Table ~\ref{table:challenging_set}). These results indicate the viability of the proposed algorithm in more general task distributions, demonstrating enhanced performance in environments that require sample efficiency with long-term credit assignment which an episodic trace is designed to solve.

\section{Discussion}

This paper deals with the problem of long-tailed distributions in reinforcement learning and is inspired by the theory of complementary systems, which states that an intelligent agent requires a fast and slow learning system acting complementary to each other. Here, the momentum loss of contrastive learning provides a mechanism to detect important long tail samples in an unsupervised manner. These samples are then prioritized in a separate buffer that also stores the hidden activation associated with such states. When a rare sample is detected, a similarity search is used to find relevant keys, and the corresponding hidden activations are merged and reinstated in the recurrent layers. The modular architecture and its improved performance in general environments like Atari suggest that this can be seamlessly integrated into other RL architectures.

This architecture relates to how the hippocampus, which is a fast learning system, acts in tandem with the slow learning cortex of the organism to store relevant experiences and replay them to overcome the statistics of the environment that the organism is subjected to~\citep{o2010play}. The episodic memory relies on the network to discover long-tail data from the incoming data stream. The network relies on the episodic memory for identifying the relevant memories from long tail data in order to reinstate it in the recurrent layers of the working memory system to execute the episodic sequence. Similarly, the brain could reinstate episodic sequences from the hippocampus to the working memory when animals execute a task.

Furthermore, dopamine neurotransmitter has been found to detect novel states and relay them to the hippocampus~\citep{duszkiewicz2019novelty} and is also related to curiosity and learning progress which is analogous to momentum loss here~\citep{ten2021humans, gruber2019curiosity}. Future work could look at how to extend this to more realistic 3D environments and also applied RL problem scenarios such as Sepsis, Road accident trajectories, etc that require a mechanism for handling rare but valuable states~\citep{fodor1988connectionism}.

\section{Conclusion}

This paper attempts to overcome the problem of long-tailed data for reinforcement learning which traditional architectures do not address well owing to their underlying assumptions. An unsupervised long-tail discovery method using self-supervised momentum loss is used to prioritize long-tail data. Using this prioritization, an episodic storing of hidden activations is done to be later reinstated in the recurrent layers so that rare trajectories are executed. Both of these proposed features are crucial in enabling the network to perform better than conventional architectures on a long-tail dataset. We hope that this work will encourage the development of new RL methods in such data distributions and finally enable the development of agents capable of learning from a lifetime of non-uniform experience.



\bibliographystyle{named}
\bibliography{ijcai24}

\newpage
\appendix
\onecolumn

\section{Appendix}

\subsection{Additional Results} \label{app:addresults}

We compare our method with two more architectures with very different methods of solving and memory. They are:
\begin{itemize}
  \item Variational Autoencoder (VAE) ~\citep{pu2016variational}
  \item Hierarchical Chunk Attention Memory (HCAM) ~\cite{lampinen2021towards}
\end{itemize}

The results on all three tasks are provided in Table ~\ref{table:performance_supp}. We notice that both these methods perform poorly compared to our method and even the baseline IMPALA. Also, these methods perform worse when tested on the rare 20\% of the tasks.

\begin{table*}[htb]
	\caption{\textbf{Evaluation Performance:} We additionally compare our method with different varying types of methods namely Variational Autoencoder (VAE) and Hierarchical Chunk Attention Memory (HCAM).}\label{table:performance_supp}
	\vskip 0.15in
    \begin{minipage}{\linewidth}
		\resizebox{\linewidth}{!}{
		\begin{tabular}{x{0.441\linewidth}x{0.147\linewidth}x{0.147\linewidth}x{0.147\linewidth}}
			\toprule
            \multirow{2}{*}{\parbox{3.5cm}{\centering \phantom{} \\ \phantom{} \\ Method}} & \multicolumn{3}{c}{Accuracy (\%) $\mid$ \textbf{Zipf's Gridworld}}                                                          \\ \cmidrule(lr){2-4} 
										& \multicolumn{1}{c}{Zipfian} & \multicolumn{1}{c}{\parbox{3.5cm}{\centering Uniform\\(All maps and objects)}} & \multicolumn{1}{c}{\parbox{3.5cm}{\centering Rare\\(Rarest 20\% objects\\on rarest 20\% maps)}}   \\ \midrule
VAE                                   &  45.3 $\pm$ 5.6 & \parbox{3.5cm}{\centering 5.7 $\pm$ 4.5} &  \parbox{3.5cm}{\centering 0.0 $\pm$ 0.0} \\
HCAM                                  &  75.9 $\pm$ 1.1 & \parbox{3.5cm}{\centering 25.6 $\pm$ 0.9} &  \parbox{3.5cm}{\centering 0.0 $\pm$ 0.0} \\
\textbf{Ours}                         &  \textbf{98.5 $\pm$ 1.2} & \parbox{3.5cm}{\centering \textbf{66.3 $\pm$ 1.0}} & \parbox{3.5cm}{\centering \textbf{25.2 $\pm$ 1.1}} \\
			\bottomrule
        \end{tabular}
		}
	\end{minipage}\\[12pt]

    \begin{minipage}{\linewidth}
		\resizebox{\linewidth}{!}{
		\begin{tabular}{x{0.441\linewidth}x{0.147\linewidth}x{0.147\linewidth}x{0.147\linewidth}}
			\toprule
            \multirow{2}{*}{\parbox{3.5cm}{\centering \phantom{} \\ \phantom{} \\ Method}} & \multicolumn{3}{c}{Accuracy (\%) $\mid$ \textbf{Zipf's 3DWorld}}                                                          \\ \cmidrule(lr){2-4} 
										& \multicolumn{1}{c}{Zipfian} & \multicolumn{1}{c}{\parbox{3.5cm}{\centering Uniform\\(All maps and objects)}} & \multicolumn{1}{c}{\parbox{3.5cm}{\centering Rare\\(Rarest 20\% objects\\on rarest 20\% maps)}}   \\ \midrule
VAE                                   &  28.3 $\pm$ 4.3 & \parbox{3.5cm}{\centering 5.7 $\pm$ 3.2} &  \parbox{3.5cm}{\centering 0.0 $\pm$ 0.0} \\
HCAM                                  &  80.1 $\pm$ 3.6 & \parbox{3.5cm}{\centering 31.6 $\pm$ 6.5} &  \parbox{3.5cm}{\centering 3.1 $\pm$ 1.2} \\
\textbf{Ours}                         &  \textbf{99.2 $\pm$ 1.3} & \parbox{3.5cm}{\centering \textbf{80.1 $\pm$ 2.0}} & \parbox{3.5cm}{\centering \textbf{55.6 $\pm$ 2.4}} \\
			\bottomrule
        \end{tabular}
		}
	\end{minipage}\\[12pt]

    \begin{minipage}{\linewidth}
            \renewcommand{\arraystretch}{1.4}
		\resizebox{\linewidth}{!}{
		\begin{tabular}{x{0.35\linewidth}x{0.10\linewidth}x{0.10\linewidth}x{0.10\linewidth}x{0.10\linewidth}x{0.10\linewidth}x{0.10\linewidth}}
			\toprule
            \multirow{2}{*}{\parbox{3.5cm}{\centering \phantom{} \\ \phantom{} \\ \phantom{} \\ Method}} & \multicolumn{6}{c}{Accuracy (\%) $\mid$ \textbf{Zipf's Labyrinth}} \\ \cmidrule(lr){2-7}
                                            & \multicolumn{3}{c}{Forward Zipf} & \multicolumn{3}{c}{Reversed Zipf} \\ \cmidrule(lr){2-4} \cmidrule(lr){5-7}
										& \multicolumn{1}{c}{Zipfian}	& \multicolumn{1}{c}{\parbox{3.5cm}{\centering Uniform\\(All maps and objects)}}	& \multicolumn{1}{c}{\parbox{3.5cm}{\centering Rare\\(Rarest 20\% objects\\on rarest 20\% maps)}} & \multicolumn{1}{c}{\parbox{3.5cm}{\centering Zipfian}}	& \multicolumn{1}{c}{\parbox{3.5cm}{\centering Uniform\\(All maps and objects)}}	& \multicolumn{1}{c}{\parbox{3.5cm}{\centering Rare\\(Rarest 20\% objects\\on rarest 20\% maps)}}   \\ \midrule
VAE                                &  35.0 $\pm$ 6.8 & \parbox{3.5cm}{\centering 3.6 $\pm$ 2.6} &  \parbox{3.5cm}{\centering 0.0 $\pm$ 0.0} &  \parbox{3.5cm}{\centering 36.2 $\pm$ 6.4} &  \parbox{3.5cm}{\centering 5.3 $\pm$ 3.2} &  \parbox{3.5cm}{\centering 0.0 $\pm$ 0.0} \\
HCAM                                &  53.1 $\pm$ 2.3 & \parbox{3.5cm}{\centering 20.3 $\pm$ 3.4} &  \parbox{3.5cm}{\centering 0.0 $\pm$ 0.0} &  \parbox{3.5cm}{\centering 56.2 $\pm$ 4.3} &  \parbox{3.5cm}{\centering 25.6 $\pm$ 2.5} &  \parbox{3.5cm}{\centering 0.0 $\pm$ 0.0} \\
\textbf{Ours}                                &  \textbf{71.3 $\pm$ 3.1} & \parbox{3.5cm}{\centering \textbf{54.2 $\pm$ 2.2}} &  \parbox{3.5cm}{\centering \textbf{32.2 $\pm$ 2.1}} &  \parbox{3.5cm}{\centering \textbf{77.5 $\pm$ 7.0}} &  \parbox{3.5cm}{\centering \textbf{47.1 $\pm$ 2.3}} &  \parbox{3.5cm}{\centering \textbf{25.3 $\pm$ 1.9}} \\
			\bottomrule
        \end{tabular}
		}
	\end{minipage}\\
\vskip -0.1in
\end{table*}

\subsection{Supplementary Analyses} \label{app:supplementary}

\begin{figure*}[ht!]
\centering
\begin{tabular}{l}
\begin{minipage}[b]{1.0 \linewidth}
\begin{center}
\subfigure[Forward Zipf]{
\includegraphics[width=3.0in]{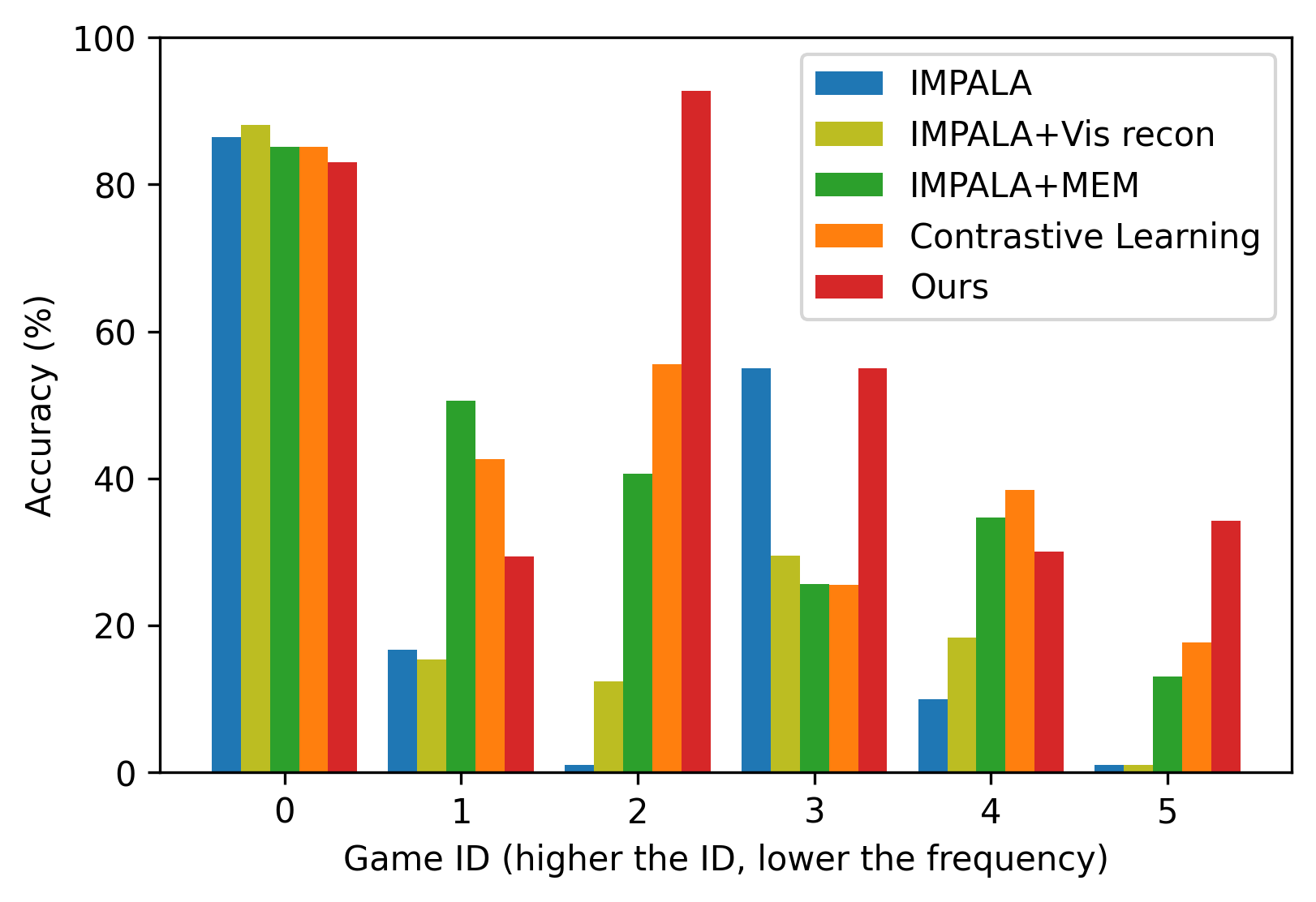}
\label{fig:sup2a}
}
\hspace{0.2cm}
\subfigure[Reversed Zipf]{
\includegraphics[width=3.0in]{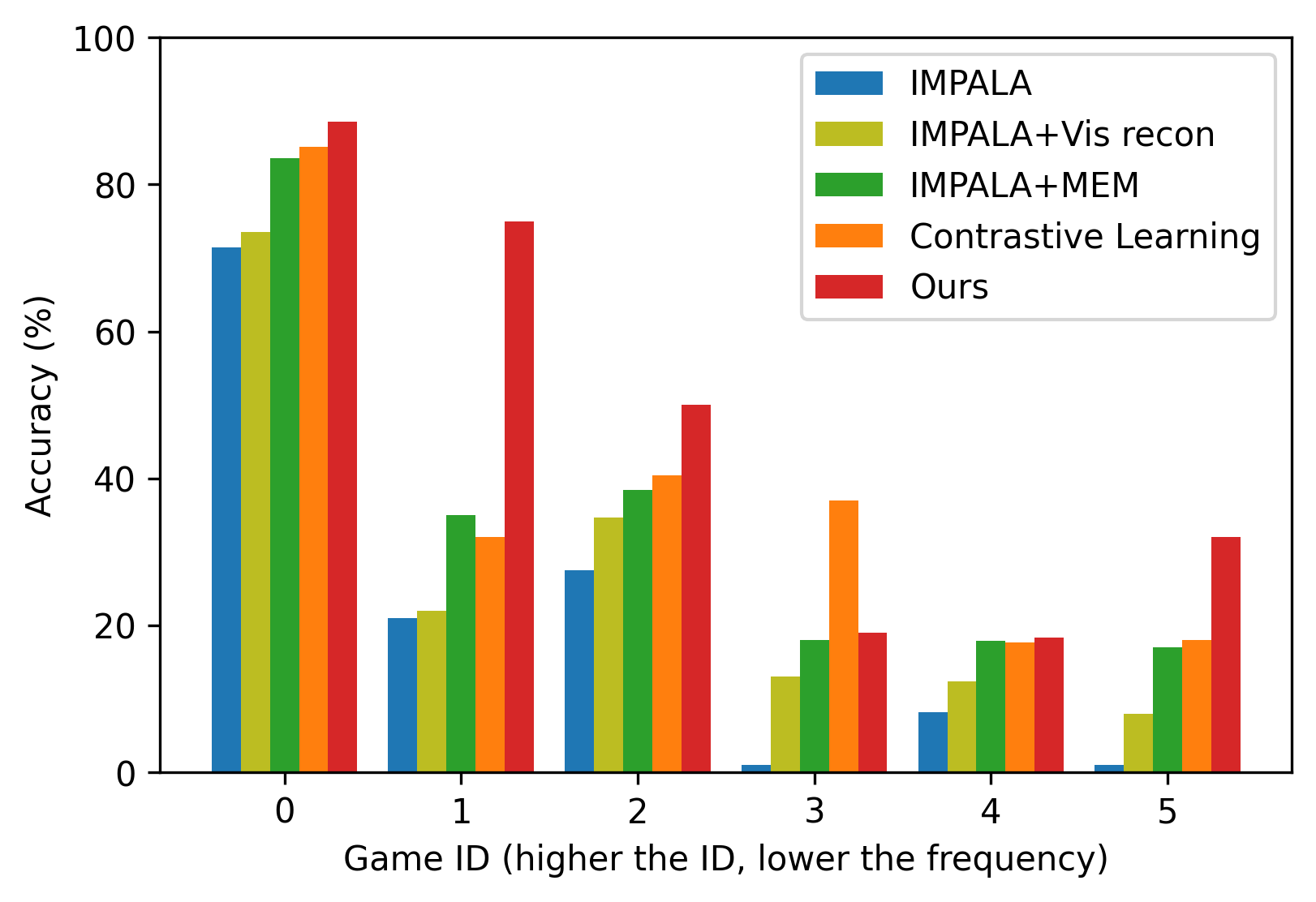}
\label{fig:sup2b}
}
\end{center}
\end{minipage}
\end{tabular}
\vspace{-0.5cm}
\caption{\textbf{Performance Plots (Zipf's Labyrinth):} Performance of agents in Zipfs Labyrinth Foward \& Reversed on all tasks.}
\label{fig:sup2}
\end{figure*}

For Zipf's Labyrinth task, we plot the accuracies for each of the methods on all tasks in Figure ~\ref{fig:sup2}. We can clearly see that our method (red bar) performs consistently well on most of the rare and medium rare trials. On the extremely rare trials (tasks), our method is the best across all architectures which shows the effectiveness of discovering long-tailed states using our proposed method.

\begin{figure}[ht]
\begin{center}
\includegraphics[width=0.65\textwidth]{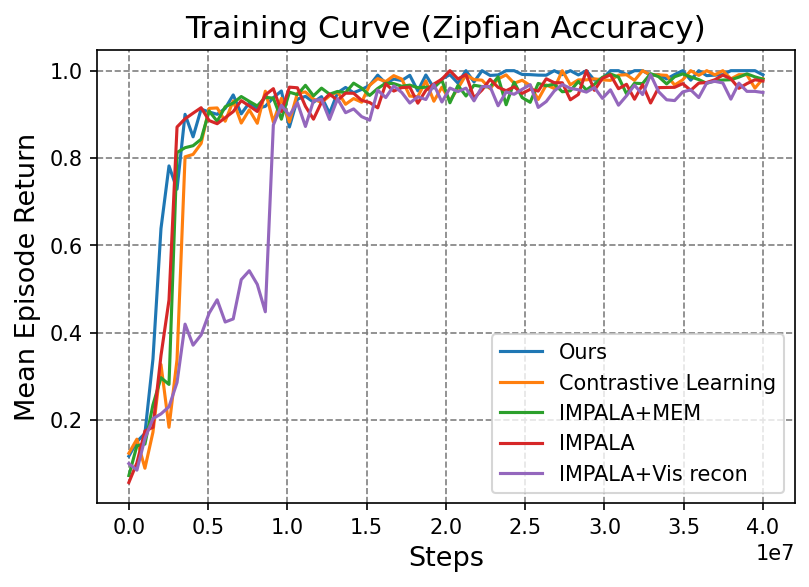}
\end{center}
\caption{\textbf{Performance plots:} Training curves for different experiments.}\label{app:maincurves}
\end{figure}

The training curve for the main experiments (Zipf's Gridworld) can be seen in Figure~\ref{app:maincurves}.

For the Zip'f Gridworld task, we only consider 10 maps and 10 objects. In the paper where the task is introduced, they have considered 20 maps and 20 objects. Reducing the number of maps and objects doesn't make the task easier, but might be more difficult because we train the agent using just 50 actors and for 4e7 steps due to computational constraints. We were unable to reproduce the exact results on 20 maps and 20 objects using our implementation of IMPALA. The original code for training on Deepmind's environments is also not publicly available yet due to some issues on their side. For similar reasons we only consider 6 tasks in Zipf's Labyrinth environment. The 6 tasks in order are listed below:

\begin{lstlisting}[language=Python,
                   numbers=left,
                   numbersep=5pt,
                   gobble=0,
                   frame=lines,
                   framesep=2mm]
LEVELS_DMLAB30_FORWARD = [
    'rooms_collect_good_objects_train', 'rooms_exploit_deferred_effects_train',
    'rooms_select_nonmatching_object', 'explore_object_locations_small',
    'explore_goal_locations_small', 'explore_object_rewards_few',
]
\end{lstlisting}

In IMPALA a subset of each trajectory is actually sent to the buffer (states at a hop of $hp$ from the start of the trajectory). For a trajectory $\mathrm{Tr} = (s_1, s_2, ..., s_k)$ of size $k$ consisting of states $s_i$, the subset of trajectory is defined as 
\begin{equation}\label{eqn:subset}
\mathrm{Tr}_{\mathrm{subset}} = \left(s_1, s_{(1+hp)}, ..., s_{\left(1+\lfloor\frac{k-1}{hp}\rfloor*hp\right)}\right)
\end{equation}

\begin{figure*}[ht!]
\centering
\begin{tabular}{l}
\begin{minipage}[b]{1.0 \linewidth}
\begin{center}
\subfigure[Effect of KNN `K' value]{
\makebox[0.5\linewidth][c]{
\includegraphics[width=7cm]{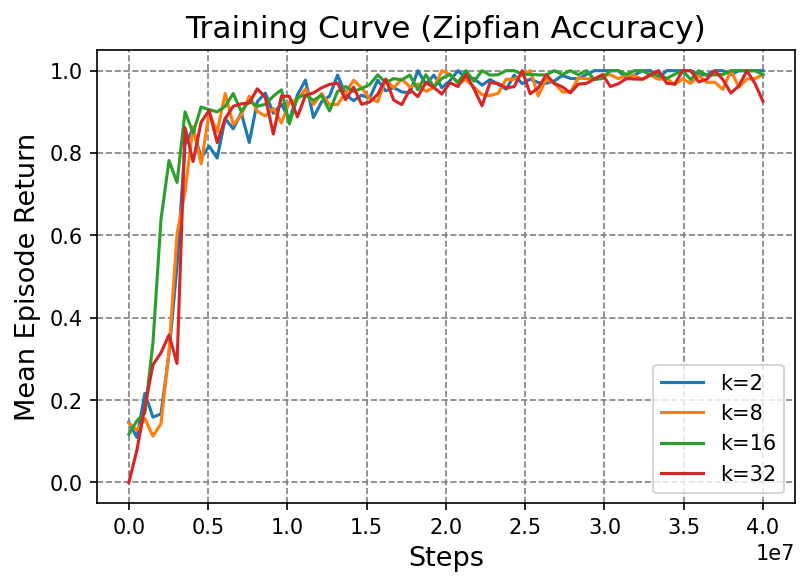}}
\label{fig:sup1a}
}
\subfigure[Effect of trajectory hop $(hp)$]{
\includegraphics[width=7cm]{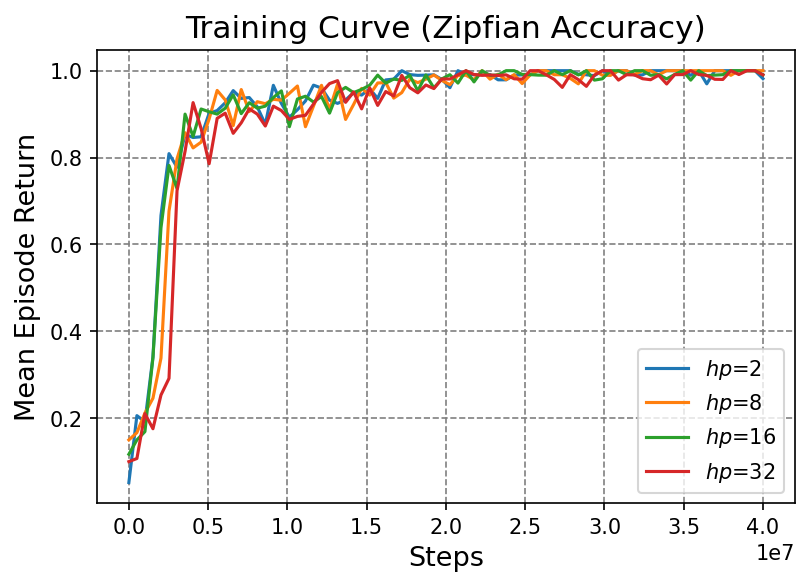}
\label{fig:sup1b}
}
\end{center}
\end{minipage}\\
\begin{minipage}[b]{1.0 \linewidth}
\begin{center}
\subfigure[Effect of transfer amount $(t_k)$]{
\makebox[0.5\linewidth][c]{
\includegraphics[width=7cm]{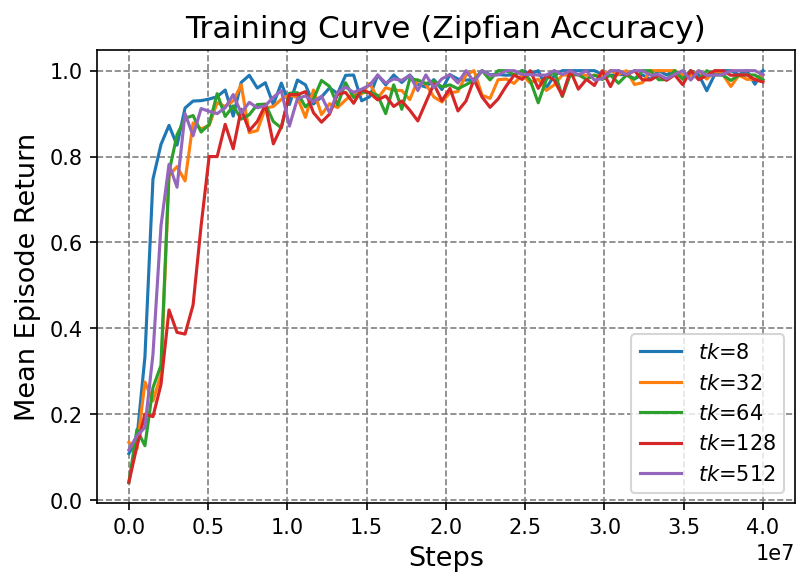}}
\label{fig:sup1c}
}
\subfigure[Effect of transfer frequency $(t_f)$]{
\includegraphics[width=7cm]{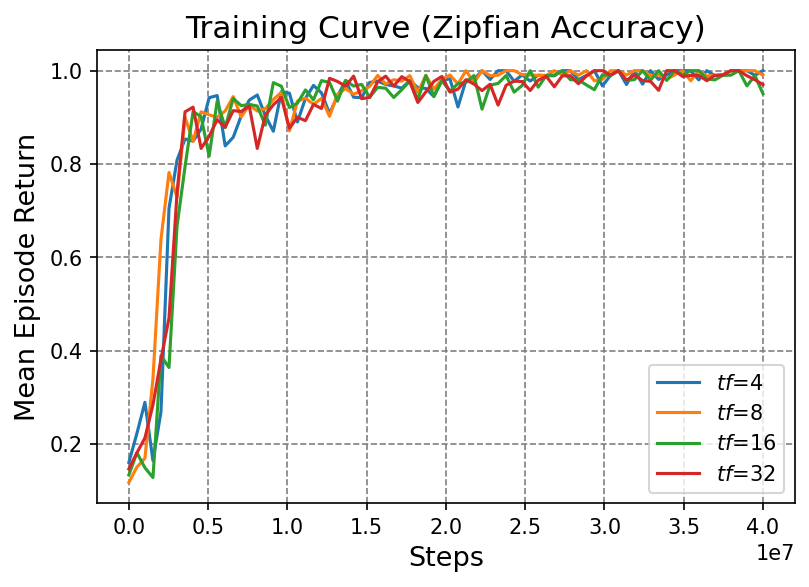}
\label{fig:sup1d}
}
\end{center}
\end{minipage}
\end{tabular}
\caption{\textbf{Training Performance:} Effect of changing different hyperparameters on training.}
\label{fig:sup1}
\end{figure*}

The effects of changing different hyperparameters of our architecture are explained here. We look at the `K' value of K Nearest Neighbour, Trajectory Hop $(hp)$, Rare State Transfer amount $(t_k)$, and Rare State Transfer Frequency $(t_f)$ for our ablation study. The effect on the training of these hyperparameters can be seen in Figure~\ref{fig:sup1}. Tables~\ref{table:subt1}-~\ref{table:subt4} give more insights into the effect of different hyperparameters on the training. Section~\ref{app:hyperparameters} gives the list of hyperparameter settings used in various experiments. 

\begin{table}[H]
\centering
\renewcommand*{\arraystretch}{1.4}
\caption{Effect of KNN `K' value.}
\vspace{0.1cm}
\begin{tabular}{|c|c|c|c|}
\hline
& \multicolumn{3}{c|}{Accuracy $(\%)$} \\ \cline{2-4}
\multicolumn{1}{|c|}{K Value} & Zipfian & \makecell{Uniform\\(All maps and objects)} & \makecell{Rare\\(Rarest 20\% objects\\on rarest 20\% maps)} \\
\hline
2                                 &  99.4 & 79.0 & 0     \\
8                                 &  98.5 & 67.6 & 0     \\
16                                &  99.9 & 84.4 & 49.9  \\
32                                &  97.3 & 74.9 & 24.9  \\
\hline
\end{tabular}
\label{table:subt1}
\end{table} 

\begin{table}[H]
\centering
\renewcommand*{\arraystretch}{1.4}
\caption{Effect of trajectory hop $(hp)$.}
\begin{tabular}{|c|c|c|c|}
\hline
& \multicolumn{3}{c|}{Accuracy $(\%)$} \\ \cline{2-4}
\multicolumn{1}{|c|}{$(hp)$} & Zipfian & \makecell{Uniform\\(All maps and objects)} & \makecell{Rare\\(Rarest 20\% objects\\on rarest 20\% maps)} \\
\hline
2                                 &  99.1 & 78.4 & 24.8 \\
8                                 &  99.3 & 79.1 & 24.9 \\
16                                &  99.9 & 84.4 & 49.9 \\
32                                &  99.1 & 77.7 & 25.2 \\
\hline
\end{tabular}
\label{table:subt2}
\end{table} 

\begin{table}[H]
\centering
\renewcommand*{\arraystretch}{1.4}
\caption{Effect of transfer amount $(t_k)$.}
\begin{tabular}{|c|c|c|c|}
\hline
& \multicolumn{3}{c|}{Accuracy $(\%)$} \\ \cline{2-4}
\multicolumn{1}{|c|}{$(t_k)$} & Zipfian & \makecell{Uniform\\(All maps and objects)} & \makecell{Rare\\(Rarest 20\% objects\\on rarest 20\% maps)} \\
\hline
8                                 &  99.2 & 81.1 & 24.9 \\
16                                &  99.9 & 84.4 & 49.8 \\
32                                &  99.2 & 76.0 & 0  \\
64                                &  98.9 & 77.9 & 0  \\
128                               &  98.7 & 69.8 & 0  \\
\hline
\end{tabular}
\label{table:subt3}
\end{table} 

\begin{table}[H]
\centering
\renewcommand*{\arraystretch}{1.4}
\caption{Effect of transfer frequency $(t_f)$.}
\begin{tabular}{|c|c|c|c|}
\hline
& \multicolumn{3}{c|}{Accuracy $(\%)$} \\ \cline{2-4}
\multicolumn{1}{|c|}{$(t_f)$} & Zipfian & \makecell{Uniform\\(All maps and objects)} & \makecell{Rare\\(Rarest 20\% objects\\on rarest 20\% maps)} \\
\hline
4                               &  99.1 & 79.9 & 25.0 \\
8                               &  99.9 & 84.4 & 49.8 \\
16                              &  99.3 & 82.0 & 49.9 \\
32                              &  99.1 & 76.9 & 74.8 \\
\hline
\end{tabular}
\label{table:subt4}
\end{table} 

We also run our experiments on the Atari Learning Environment ~\citep{bellemare2013arcade}, specifically across the set of 57 Atari games previously examined by ~\citep{DBLP:journals/corr/SchaulQAS15}. This collection of games presents a compelling array of tasks characterized by multifaceted challenges, encompassing aspects such as sparse reward structures and considerable variations in scoring scales across the diverse game set. The results shown in Table~\ref{table:atari_individual_games} indicate that the proposed method outperforms both variations of IMPALA on 32 out of the 56 tasks considered (57.14\%). Further, on a subset of Atari games namely,~\emph{challenging set}~\citep{pmlr-v119-badia20a}, we get better results on 7/10 tasks that emphasize hard exploration with long-term credit assignment  (Table ~\ref{table:challenging_set}). 

\begin{table}[H]
\centering
\renewcommand*{\arraystretch}{1.4}
\caption{Results on Atari games \emph{challenging set} of~\cite{pmlr-v119-badia20a}.}
\begin{tabular}{|c|c|c|c|}
\hline
\multicolumn{1}{|c|}{Environment} & IMPALA (shallow) & \makecell{IMPALA (deep)} & \makecell{\textbf{Ours}} \\
\hline
Beam Rider & 8219.92 & 32463.47 & \textbf{44938.83} \\
Freeway & 0.00 & 0.00 & \textbf{0.33} \\
Gravitar & 211.50 & 359.50 & \textbf{419.64} \\
Montezuma's & 0.00 & 0.00 & \textbf{86.34} \\
Pitfall & -11.14 & -1.66 & -15.35 \\
Pong & 20.40 & 20.98 & 20.58 \\
Private Eye & 92.42 & 98.50 & \textbf{99.64} \\
Skiing & -29975.00 & -10180.38 & -29999.24 \\
Solaris & 2368.40 & 2365.00 & \textbf{2483.64} \\
Venture & 0.00 & 0.00 & \textbf{0.00} \\
\hline
\end{tabular}
\label{table:challenging_set}
\end{table} 

\begin{center}
  \small
  \captionof{table}{Atari scores after training for 200M steps in environment. Existing results taken from ~\citep{espeholt2018impala}.}
  \begin{tabular}{l@{\hspace{.26cm}}r@{\hspace{.26cm}}r@{\hspace{.26cm}}r@{\hspace{.26cm}}r@{\hspace{.26cm}}r@{\hspace{.26cm}}}
  \toprule
                             &         IMPALA (shallow) &         IMPALA (deep) &         \textbf{Ours} \\
  \midrule
          alien              &         1536.05          &         15962.10       & \textbf{21948.24}     \\
          amidar             &         497.62           & \textbf{1554.79}       &         324.34        \\
          assault            &         12086.86         &         19148.47       & \textbf{25142.24}     \\
          asterix            &         29692.50         &         300732.00      & \textbf{382749.53}    \\
          asteroids          &         3508.10          &         108590.05      & \textbf{128394.94}    \\
          atlantis           &         773355.50        &         849967.50      & \textbf{927883.39}    \\
          bank\_heist        &         1200.35          & \textbf{1223.15}       &         1193.34       \\
          battle\_zone       &         13015.00         & \textbf{20885.00}      &         19238.34      \\
          beam\_rider        &         8219.92          &         32463.47       & \textbf{44938.83}     \\
          berzerk            &         888.30           &         1852.70        & \textbf{1928.46}      \\
          bowling            &         35.73            & \textbf{59.92}         &         20.53         \\
          boxing             &         96.30            & \textbf{99.96}         &         97.34         \\
          breakout           &         640.43           &         787.34         & \textbf{928.64}       \\
          centipede          &         5528.13          &         11049.75       & \textbf{28183.64}     \\
          chopper\_command   &         5012.00          &         28255.00       & \textbf{28442.89}     \\
          crazy\_climber     &         136211.50        & \textbf{136950.00}     &         136212.7      \\
          defender           &         58718.25         &         185203.00      & \textbf{192837.78}    \\
          demon\_attack      &         107264.73        & \textbf{132826.98}     &         135454.52     \\
          double\_dunk       &         -0.35            & \textbf{-0.33}         &         -37           \\
          enduro             &         0.00             &         0.00           & \textbf{0.00}         \\
          fishing\_derby     &         32.08            & \textbf{44.85}         &         42.45         \\
          freeway            &         0.00             &         0.00           & \textbf{0.33}         \\
          frostbite          &         269.65           & \textbf{317.75}        &         310.23        \\
          gopher             &         1002.40          &         66782.30       & \textbf{62838.46}     \\
          gravitar           &         211.50           &         359.50         & \textbf{419.64}       \\
          hero               &         33853.15         &         33730.55       & \textbf{33854.35}     \\
          ice\_hockey        &         -5.25            & \textbf{3.48}          &         3.21          \\
          jamesbond          &         440.00           &         601.50         & \textbf{728.35}       \\
          kangaroo           &         47.00            & \textbf{1632.00}       &         1536.64       \\
          krull              &         9247.60          &         8147.40        & \textbf{10293.93}     \\
          kung\_fu\_master   &         42259.00         & \textbf{43375.50}      &         42474.4       \\
          montezuma\_revenge &         0.00             &         0.00           & \textbf{86.34}        \\
          ms\_pacman         &         6501.71          & \textbf{7342.32}       &         7293.43       \\
          name\_this\_game   &         6049.55          &         21537.20       & \textbf{23847.83}     \\
          phoenix            &         33068.15         & \textbf{210996.45}     &         208494.24     \\
          pitfall            &         -11.14           & \textbf{-1.66}         &         -15.35        \\
          pong               &         20.40            & \textbf{20.98}         &         20.58         \\
          private\_eye       &         92.42            &         98.50          & \textbf{99.64}        \\
          qbert              &         18901.25         &         351200.12      & \textbf{362748.22}    \\
          riverraid          &         17401.90         & \textbf{29608.05}      &         28793.53      \\
          road\_runner       &         37505.00         &         57121.00       & \textbf{59304.52}     \\
          robotank           &         2.30             & \textbf{12.96}         &         1.34          \\
          seaquest           &         1716.90          & \textbf{1753.20}       &         1743.64       \\
          skiing             &         -29975.00        & \textbf{-10180.38}     &         -29999.24     \\
          solaris            &         2368.40          &         2365.00        & \textbf{2483.64}      \\
          space\_invaders    &         1726.28          &         43595.78       & \textbf{46383.91}     \\
          star\_gunner       &         69139.00         &         200625.00      & \textbf{218373.24}    \\
          tennis             &         -1.89            & \textbf{0.55}          &         -2.44         \\
          time\_pilot        &         6617.50          &         48481.50       & \textbf{50283.15}     \\
          tutankham          &         267.82           &         292.11         & \textbf{299.34}       \\
          up\_n\_down        &         273058.10        & \textbf{332546.75}     &         315554.39     \\
          venture            &         0.00             &         0.00           & \textbf{0.00}         \\
          video\_pinball     &         228642.52        &         572898.27      & \textbf{603947.23}    \\
          wizard\_of\_wor    &         4203.00          & \textbf{9157.50}       &         8923.42       \\
          yars\_revenge      &         80530.13         &         84231.14       & \textbf{85039.92}     \\
          zaxxon             &         1148.50          &         32935.50       & \textbf{34923.10}     \\
  \bottomrule
  \label{table:atari_individual_games}
  \end{tabular}
  \end{center}

\subsection{Experiment Hyperparameters} \label{app:hyperparameters}

$
\begin{array}{lrrr}
\hline & \text { Zipf's Gridworld } & \text { Zipf's 3DWorld } & \text { Zipf's Labyrinth } \\
\hline \text { Image Width } & 84  & 84  & 84 \\
\text { Image Height } & 84  & 84  & 84 \\
\text { Action Repeats } & 1 & 3 & 1 \\
\text { Unroll Length } & 32 & 32 & 32 \\
\text { Discount }(\gamma) & 0.99 & 0.99 & 0.99 \\
\text { Baseline loss scaling } & 0.5 & 0.6 & 0.5 \\
\text { Entropy cost } & 0.01 & 0.00001 & 0.01 \\
\text { Optimizer } & \text { RMSProp } & \text { RMSProp } & \text { RMSProp } \\
\text { Learning rate} & 3e-4 & 3e-4 & 3e-4 \\
\text { Number of training steps } & 4e7 & 4e7 & 4e7 \\
\text { Maximum steps in a trial } & 100 & 200 & 500 \\
\hline
\end{array}
$

$\\\\
\begin{array}{lrrr}
\hline & \text { Additional Parameters } & \text { Zipf's 3DWorld } & \text { Zipf's Labyrinth } \\
\hline \text { Zipf's Exponent $(e)$ } & 2  & 2  & 2 \\
\text { Number of Actors } & 50 & 50 & 50 \\
\text { Trajectory Hop $(hp)$ } & 16 & 16 & 16 \\
\text { Average Momentum Beta $(\beta)$ } & 0.97 & 0.97 & 0.97 \\
\text { Loss Gamma $(\gamma)$ } & 0.5 & 0.5 & 0.5 \\
\text { MEM Buffer capacity } & 1024 & 2048 & 2048 \\
\text { Familiarity Memory Buffer capacity } & 1024 & 2048 & 2048 \\
\text { Rare State Transfer Amount $(t_k)$ } & 512 & 512 & 512 \\
\text { Rare State Transfer Frequency $(t_f)$ } & 8 & 8 & 8 \\
\text { KNN $(K)$ } & 16 & 32 & 32 \\
\text { Epsilon $(\epsilon)$ } & 1e-3 & 1e-3 & 1e-3 \\
\text { Sigma $(\sigma)$ } & 0.05 & 0.05 & 0.05 \\
\hline
\end{array}
$


\end{document}